%% file: main.tex
\documentclass[sigconf,natbib=true,anonymous=false,authorversion]{acmart}

\usepackage{multirow}

\AtBeginDocument{%
  \providecommand\BibTeX{{%
    \normalfont B\kern-0.5em{\scshape i\kern-0.25em b}\kern-0.8em\TeX}}}

\copyrightyear{2022}
\acmYear{2022}
\setcopyright{acmlicensed}\acmConference[SIGIR '22]{Proceedings of the 45th International ACM SIGIR Conference on Research and Development in Information Retrieval}{July 11--15, 2022}{Madrid, Spain}
\acmBooktitle{Proceedings of the 45th International ACM SIGIR Conference on Research and Development in Information Retrieval (SIGIR '22), July 11--15, 2022, Madrid, Spain}
\acmPrice{15.00}
\acmDOI{10.1145/3477495.3531726}
\acmISBN{978-1-4503-8732-3/22/07}

\acmSubmissionID{rs1276}

\begin{document}

\input{stats}

\title[Monant Medical Misinformation Dataset]{Monant Medical Misinformation Dataset: Mapping Articles to Fact-Checked Claims}

\author{Ivan Srba}
\orcid{0000-0003-3511-5337}
\affiliation{%
  \institution{Kempelen Institute of Intelligent Technologies}
  \streetaddress{Mlynske nivy 5}
  \city{Bratislava}
  \country{Slovakia}
}
\email{ivan.srba@kinit.sk}

\author{Branislav Pecher}
\orcid{0000-0003-0344-8620}
\affiliation{%
  \institution{Faculty of Information Technology, Brno University of Technology}
  \city{Brno}
  \country{Czech Republic}
}
\additionalaffiliation{
  \institution{Kempelen Institute of Intelligent Technologies}
  \streetaddress{Mlynske nivy 5}
  \city{Bratislava}
  \country{Slovakia}
}
\email{branislav.pecher@kinit.sk}

\author{Matus Tomlein}
\orcid{0000-0002-9960-700X}
\affiliation{
  \institution{Kempelen Institute of Intelligent Technologies}
  \streetaddress{Mlynske nivy 5}
  \city{Bratislava}
  \country{Slovakia}
}
\email{matus.tomlein@kinit.sk}

\author{Robert Moro}
\orcid{0000-0002-3052-8290}
\affiliation{
  \institution{Kempelen Institute of Intelligent Technologies}
  \streetaddress{Mlynske nivy 5}
  \city{Bratislava}
  \country{Slovakia}
}
\email{robert.moro@kinit.sk}

\author{Elena Stefancova}
\orcid{0000-0001-8683-939X}
\affiliation{
  \institution{Kempelen Institute of Intelligent Technologies}
  \streetaddress{Mlynske nivy 5}
  \city{Bratislava}
  \country{Slovakia}
}
\email{elena.stefancova@kinit.sk}

\author{Jakub Simko}
\orcid{0000-0003-0239-4237}
\affiliation{
  \institution{Kempelen Institute of Intelligent Technologies}
  \streetaddress{Mlynske nivy 5}
  \city{Bratislava}
  \country{Slovakia}
}
\email{jakub.simko@kinit.sk}

\author{Maria Bielikova}
\orcid{0000-0003-4105-3494}
\affiliation{
  \institution{Kempelen Institute of Intelligent Technologies}
  \streetaddress{Mlynske nivy 5}
  \city{Bratislava}
  \country{Slovakia}
}
\additionalaffiliation{
  \institution{slovak.AI}
  \streetaddress{Ilkovicova 2}
  \city{Bratislava}
  \country{Slovakia}
}
\email{maria.bielikova@kinit.sk}

\renewcommand{\shortauthors}{Srba et al.}

\begin{abstract}
   False information has a significant negative influence on individuals as well as on the whole society. Especially in the current COVID-19 era, we witness an unprecedented growth of medical misinformation. To help tackle this problem with machine learning approaches, we are publishing a feature-rich dataset of approx. {\numberOfArticlesApprox} medical news articles/blogs and {\numberOfClaimsApprox} fact-checked claims. It also contains {\numberOfManualLabels} manually and more than {\numberOfClaimMappingsApprox} automatically labelled mappings between claims and articles. Mappings consist of claim presence, i.e., whether a claim is contained in a given article, and article stance towards the claim. We provide several baselines for these two tasks and evaluate them on the manually labelled part of the dataset. The dataset enables a number of additional tasks related to medical misinformation, such as misinformation characterisation studies or studies of misinformation diffusion between sources.
\end{abstract}



\maketitle

\section{Introduction}
\label{intro}
False information on the Web has been a widely researched phenomenon in computer science for the past few years, as evidenced by many recent surveys, e.g.,~\cite{shu2017fake-data-mining, bondielli2019survey, zhou2020fake-news-survey, guo2020future-of-false-info, Zubiaga2018RumoursSocialMedia, zhang2020overview}. The main focus was initially on political fake news; however, it shifted towards the medical domain with the arrival of COVID-19 pandemic and an \textit{infodemic} (a surge of new misinformation\footnote{We use the term \textit{misinformation} to describe false or misleading information that is created and spread regardless of an intention to deceive, in contrast to \textit{disinformation}, which refers specifically to false information created and spread deliberately.} related to the COVID-19).

Motivated by significant negative consequences of online false information, a number of approaches based on information retrieval and machine learning have been proposed to detect it. The main branch of existing works rely on \textit{indirect features} derived from content (textual as well as multimedia) and context, such as content style, propagation patterns, author/source credibility, or social engagements/consumption~\cite{zhou2020fake-news-survey}. This approach has several advantages, e.g., it allows to detect new cases of false information early (since new false information usually share similar characteristics with prior cases). On the other hand, the existing methods usually provide only limited single-label classification (typically a binary one -- a news article/blog/social media post is/is not a piece of false information), have insufficient explainability, and, in addition, they may suffer from domain shifts (either natural changes in domain characteristics or targeted adversarial attacks). 

Another branch of knowledge-based approaches evaluates the actual content veracity by performing a \textit{fact-checking}. Fact-checking stands for detection and verification of a claim, such as ``Drinking bleach or pure alcohol can cure the coronavirus infections''\footnote{\url{https://www.who.int/emergencies/diseases/novel-coronavirus-2019/advice-for-public/myth-busters}}, against a knowledge base (e.g., scientific articles~\cite{Hashavit2021}, articles from sources deemed reliable, such as Wikipedia~\cite{thorne-etal-2018-fever}, or knowledge graphs of known facts~\cite{zhong-etal-2020-reasoning,zhao2020transformer}). This approach may be preferable in many situations, including tackling false information in medical domain that (from its inherent characteristics) requires accurate, easily explainable and robust approaches for misinformation detection.

Fact-checking can be done either manually by professional fact-checkers or (semi-)automatically with the help of AI. Manual fact-checking is time consuming and, yet, scale-insufficient. On the other hand, fully automatic end-to-end fact-checking (e.g.,~\cite{Hassan2017ClaimBuster}) is a challenging task and existing solutions have not yet achieved a sufficient accuracy, generality, and credibility~\cite{nakov_automated_2021}. The real promise of technologies for now lies in tools to assist fact-checkers to identify and investigate claims, and to deliver their conclusions as effectively as possible~\cite{Graves2018UnderstandingFactChecking,nakov_automated_2021}.

AI research may assist fact-checkers in the following steps of the fact-checking process~\cite{nakov_automated_2021,zeng_automated_2021}: 1) identification of claims worth fact-checking, 2) detection of previously fact-checked claims relevant to the identified fact-check-worthy claims, 3) retrieving relevant evidence to fact-check a claim, and 4) verification of the claim based on the retrieved evidence. In addition, the set of already fact-checked claims can be mapped back to additional (already existing or new) online content. While similar to step 2 above, here the input is a fact-checked claim and the output a list of articles containing the claim. Thus, it can be viewed as the fifth (dissemination) step in the fact-checking process, which is typically not done in manual fact-checking as it is difficult or even impossible for the fact-checkers to manually find/update such relevant content~\cite{Wang2018RelevantDocumentDiscovery}. Especially in medical domain, many misinformative articles reuse claims, which have already been expertly fact-checked, thus making the use of existing databases of fact-checks feasible.

AI-based fact-checking support in the multiple steps above is fundamentally based on \textit{document to claim mapping} (document being a news article/blog, a social media post, etc.) and more specifically on two IR/NLP tasks: \textit{presence detection} and \textit{stance classification}~\cite{riedel2017FNCbaseline, hanselowski2018athene, Umer2020StanceDetection, borges2019combiningFNC, Zhang2019Stance, baly2018integrating, hardalov2021survey}. The detection of previously fact-checked claims (step 2), became a target of research interest only recently~\cite{shaar2020known} and is one of the least studied research problems related to fact-checking ~\cite{nakov_automated_2021}. It is typically addressed at the claim to claim level, i.e., previously fact-checked claims are ranked based on their relevance for a single given claim~\cite{shaar2020known,mansour2022}. Nevertheless, very recently, Shaar et al.~\cite{shaar_assisting_2021} formulated a more challenging version of this task as identification of all previously fact-checked claims in an input document (that can potentially contain multiple check-worthy claims). The task includes detection of a document's sentences containing any of the previously fact-checked claims and the stance of these sentences towards the present claims. Presence detection and stance classification are also crucial for the other steps of the fact-checking process. In step 4, presence of an investigated claim is detected to retrieve evidence and then its stance towards the claim is used to verify factuality of the claim. Finally, both presence and stance are used in step 5 to map already fact-checked claims to additional documents~\cite{Wang2018RelevantDocumentDiscovery}.

While situation with datasets for the first branch of false information detection (based on content and contextual characteristics) continually improves (cf. Section~\ref{related-work}), datasets for AI research on fact-checking, particularly datasets providing a mapping between documents and claims (claim presence and document stance), still present a major problem hampering further research.

In this paper, we are introducing a novel \emph{medical misinformation dataset}. It contains:

\begin{itemize}
    \item full-texts, original source URL, and other extracted metadata of approx. {\numberOfArticlesApprox} news articles and blog posts on medical topics published between January 1, 1998 and February 1, 2022 from a total of {\numberOfArticleSources} reliable and unreliable sources;
    
    \item annotations with a source credibility score from expertly-curated lists, such as Media Bias/Fact Check, when it is available;
    
    \item around {\numberOfClaimsApprox} fact-checks and extracted verified medical claims with their unified veracity ratings published by fact-checking organisations such as Snopes or FullFact;
    
    \item {\numberOfManualLabels} manually and more than {\numberOfClaimMappingsApprox} automatically labelled mappings between previously verified claims and the articles; mappings consist of two values: \textit{claim presence} (i.e., whether a claim is contained in a given article) and \textit{article stance} (i.e., whether a given article supports or rejects a claim or provides both sides of the argument).
\end{itemize}

The dataset is primarily intended to be used as a training and evaluation set for machine learning methods for claim presence detection and article stance classification, but it enables a range of other misinformation related tasks, such as misinformation characterisation, analyses of misinformation spreading or classification of source reliability. Its novelty and our main contributions lie in:

\begin{enumerate}
    \item focus on medical news articles and blog posts as opposed to social media posts or political discussions;
    \item providing multiple modalities (beside full-texts of the articles, there are also images and videos), thus enabling research of multimodal approaches;
    \item mapping of the articles to the fact-checked claims (with manual as well as predicted labels);
    \item providing source credibility labels for {\percentageOfArticlesWithReliabilityLabel} of all articles and other potential sources of weak labels that can be mined from the articles' content and metadata.
\end{enumerate}

The dataset has been collected with our universal and extensible platform Monant~\cite{srba2019monant}, which was designed to monitor, detect, and mitigate false information. We are publishing a static dump of the dataset\footnote{A sample of the data together with accompanying documentation and analyses in Jupyter notebooks is available at \url{https://github.com/kinit-sk/medical-misinformation-dataset/}. The full static dump is available at \url{https://doi.org/10.5281/zenodo.5996864}.}. Moreover, the dataset in Monant is being continuously updated with latest articles and fact-checked claims from medical and other domains (e.g., general news) and also in languages other than English (currently in Slovak and Czech). To access the live version of the dataset, the Monant platform provides an easy-to-use access by the means of a REST API\footnote{Request for REST API access can be submitted at: \url{https://doi.org/10.5281/zenodo.5996864}}.

\section{Existing datasets}
\label{related-work}

The majority of existing datasets~\cite{Sharma2019} are created for purpose of single-label false information detection. They are commonly annotated only by some simple heuristics (e.g., the veracity of articles is determined by the credibility of their sources\footnote{\url{https://github.com/several27/FakeNewsCorpus}, \url{https://www.kaggle.com/mrisdal/fake-news}}~\cite{DBLP:journals/corr/HorneA17,Hardalov2016, hossain2020banfakenews}). However, such heuristics do not necessarily capture the real veracity of the articles (e.g., articles published by reliable sources may sometimes contain misinformative content and vice versa) and, therefore, should be used only as weak labels. Contrary to that, datasets annotated manually remain small or not fully annotated (e.g., just by its title~\cite{wang2020weak}).

Another way to create single-label fake news datasets is to take advantage of fact-checks -- by following a direct link from fact-checking articles to debunked online content (news article/blogs, social media posts, etc.). Examples of such datasets are FakeNewsNet~\cite{shu2020fakenewsnet} (rich dataset providing social context from Twitter), FakeCovid dataset~\cite{shahi2020} (providing 5,182 fact-checking articles related to COVID-19 circulated in 105 countries from 92 fact-checkers, however, without the debunked content itself), CoAID dataset~\cite{cui2020coaid} (providing the mapping of fact-checking articles to debunked content, although the number of news articles covered by the dataset is quite small), or FakeHealth dataset~\cite{dai2020ginger} (providing expertly annotated news stories published at HealthNewsReview.org\footnote{\url{https://www.healthnewsreview.org/}} together with their social engagements on Twitter). These datasets do not work explicitly with claims themselves and mostly use fact-checks just to transfer veracity label to the original content. Thus, their suitability for training AI models to support steps in the fact-checking process is limited.

Specific fact-checking datasets~\cite{guo2022survey,zeng_automated_2021} are therefore created to support individual steps of fact-checking process by researchers as well as in data challenges (most prominently at CLEF CheckThat! Lab\footnote{\url{https://sites.google.com/view/clef2022-checkthat/home}}, e.g.,~\cite{Shaar2021clef, shaar_overview_2021}). Most of them are focused on political domain (political debates) and short social media posts (mostly from Twitter~\cite{Mitra2015CREDBANKAL}, Facebook\footnote{\url{https://github.com/BuzzFeedNews/2016-10-facebook-fact-check}}, or Reddit~\cite{nakamura2020rfakeddit}). However, fact-checking datasets focused on medical domain and providing mappings between claims and larger documents (such as news articles and blogs) are generally lacking. This presents a problem, because even though social media play a significant role in creation and dissemination of medical misinformation~\cite{suarez-lledo_prevalence_2021}, many people are exposed to it also when they search online for health-related issues (which is done by 72\% of adult internet users according to Pew Research Center\footnote{\url{https://www.pewresearch.org/fact-tank/2014/01/15/the-social-life-of-health-information/}}). In~\cite{Wang2018RelevantDocumentDiscovery}, the authors created a large manually-annotated dataset (covering different domains). They mapped fact-checking articles to relevant documents containing the fact-checked claims along with stance of the documents. Unfortunately, this dataset is not public. Recently, Shaar et al.~\cite{shaar_assisting_2021} created a dataset providing presence and stance mappings between larger documents and previously fact-checked claims, nevertheless, the dataset is not available yet and it is focused on political fact-checked claims only.

We can conclude that a publicly available, feature-rich, and large enough dataset containing medical news articles/blogs with labelled mappings between articles and fact-checked claims is still missing. In contrast to the described datasets, our work specifically focuses on creating a dataset containing news articles/blogs only. Focusing on one content type allows us to extract a rich set of metadata (e.g., articles' authors, sources, categories). To achieve a large set of labelled data, we do not rely on links between fact-checking articles and news articles/blogs, which are often missing. Rather, we provide both manually human-created and automatically predicted labels of claim presence and article stance which we aggregate into article-claim pair veracities.

\section{Methodology}
\label{methodology}

\subsection{Data collection methodology}
\label{methodology:collection}
To create a medical misinformation dataset of news articles/blogs and fact-checked claims (and to continuously obtain new data), we used our research platform Monant~\cite{srba2019monant}. Scraping of the relevant web content and extraction of metadata is implemented by the means of so called \textit{monitors} and \textit{data providers}. Data providers implement the scraping functionality. General parsers (from RSS feeds, WordPress sites, Google Fact Check Tool\footnote{\url{https://toolbox.google.com/factcheck/explorer}}, etc.) as well as custom crawlers and parsers were implemented (e.g., for the fact-checking site Snopes). Monitors define which data providers should be used, their scheduling (i.e., frequency of extractions), parameters setup (e.g., a list of RSS feed URLs used as an input to the RSS feed parser), and data provider chaining (if additional data providers should be chained, e.g., when a new article is found). All data is stored in a unified format in a central data storage.

To compile a list of medical English news sites/blogs and to determine their credibility, we used expertly-curated lists of reliable and unreliable sites (e.g., Media Bias/Fact Check\footnote{\url{https://mediabiasfactcheck.com/conspiracy/}} or OpenSources\footnote{\url{https://github.com/BigMcLargeHuge/opensources}}) and previous related works (e.g.,~\cite{dhoju2019}). We added additional sources of unknown credibility that were often referenced (linked) by the sources in the initial list. Next, we checked for each source, whether it still existed and how the data could be obtained from it (e.g., using a WordPress or RSS feed parser or if it required a custom parser). We ended up with a list of {\numberOfArticleSources} medical sources in English; we have a credibility (reliability) score for {\numberOfArticleSourcesWithReliabilityAnnnotation} of them. Examples of reliable (credible) sources include healthline.com, or who.int; examples of sources marked by the listings as unreliable are naturalnews.com, or healthimpactnews.com. Most of these sources contain only medical content and thus no additional content selection was needed. If a source contained articles falling under multiple topics (e.g., politics, home news), we restricted the scraping only to a category corresponding to medical/health news/blogs. 

Next, we searched for fact-checking sources that also perform fact-checking of medical claims; we compiled a list of {\numberOfClaimSources} of them (namely Snopes.com, Meta\-Fact.io, Fact\-Check.org, Politi\-fact.com, Full\-Fact.org, Health\-Feedback.org, and Science\-Feedback.co). Similarly to the case of news sites/blogs, we either collected all fact-checking articles in the case of medical-only fact-checking sites or relied on categories manually assigned by the expert fact-checkers. Since the fact-checked claims in the selected sources are explicitly stated by the fact-checkers, it was possible to automatically extract claims from the fact-checking articles. Additional claims were supplemented from the list of unproven cancer treatments published by~\cite{ghenai2018}. As veracity ratings can differ between fact-checkers, we unified them into a scale of 6 values: false, mostly false, true, mostly true, mixture, and unknown (meaning a veracity of the claim could not be evaluated by a fact-checker or experts' consensus has not been reached yet). The latter originates mostly from the MetaFact.io site, where the experts' evaluations are crowdsourced (in comparison with other fact-checking portals where the fact-checking process is typically done by one expert only) and the claim veracity is determined only when the evaluation of a sufficient number of experts is available.

\subsection{Data labelling methodology}
\label{methodology:labelling}

Our aim was to obtain manual ground-truth labels of \emph{claim presence}, i.e., whether a given verified (fact-checked) claim is present in an article, and of \emph{article stance}, i.e., what the stance of the article is towards the matched claim. Our proposed data labelling methodology was inspired by the work of Wang et al.~\cite{Wang2018RelevantDocumentDiscovery}. The labelling is performed in four steps: First, we identify possible article-claim pairs to label. Second, the pairs are distributed to annotators in batches guaranteeing that one pair is given to multiple annotators to minimise possible mistakes in the labelling process and that the same annotator never sees the same pairs multiple times, even across batches. Next, the pairs are annotated by the annotators. Lastly, the labels from all annotators are aggregated into a single claim presence and article stance label for each labelled article-claim pair. 

A total number of 28 annotators participated in the labelling process, including the authors of this paper, master students, and other researchers. To prevent potential subjectivity and low-quality labels, a match of at least two annotators had to be achieved for the label to be included into the dataset. When there was no match between the first two annotators, the article-claim pair was assigned to up to 3 additional ones to collect more labels. Overall, inter-annotator agreement was high; additional annotator was required only in 8.57\% of cases for claim presence and in 6.94\% of cases for article stance labels. In quite rare cases, when the agreement was not reached (covering difficult to annotate or disputable cases), the article-claim pair was disregarded.

\subsubsection{Labels and their aggregation}
\label{methodology:labelling:labels}

To annotate claim presence, the annotators could select one of four possible labels:

\begin{enumerate}
    \item \textbf{Present} -- when the annotator can find a part of the article (a sentence or a paragraph) that literally or semantically contains the claim.
    \item \textbf{Suggestive} -- when the article relates to the claim, but the annotator cannot identify any specific part of the article that contains it (e.g., an article discusses the flu vaccine efficacy and suggests that they are ineffective or even harmful by providing anecdotal evidence but never explicitly makes that claim).
    \item \textbf{Not present} -- when the claim is not present in the article.
    \item \textbf{Can't tell} -- when the annotator cannot, for some reason, choose any of the options above.
\end{enumerate}

When the annotators selected either ``Present'' or ``Suggestive'' label, they were further asked to label the stance of the article towards the identified claim, by selecting one of four possible labels:

\begin{enumerate}
    \item \textbf{Supporting} -- when the article supports the claim (directly or indirectly from its context).
    \item \textbf{Contradicting} -- when the article contradicts the claim (directly or indirectly from its context).
    \item \textbf{Neutral} -- when the article does not take a stand on the claim or presents arguments both \emph{for} and \emph{against} the claim.
    \item \textbf{Can't tell} -- when the annotator cannot, for some reason, choose any of the options above.
\end{enumerate}

The individual article-claim pair labels are aggregated as follows: First, we filter out all ``Can't tell'' labels. Next, if any of the remaining claim presence or article stance labels was chosen by two or more annotators for a given article-claim pair, this label is assigned as the final aggregated one. In case of no match in claim presence labels, we lower the requirement by joining the ``Present'' and ``Suggestive'' labels into one and check again for a match. If a match is found, we assign a ``Suggestive'' label as the final aggregate claim presence label. It is also worth noting that article stance labels can be evaluated only when a given claim is present in the article. As a result, there is a lower number of article stance labels compared to the number of claim presence ones.

\subsubsection{Selection of article-claim pairs for labelling}
\label{methodology:labelling:selection}

The number of all possible article-claim pairs is equal to the number of claims times the number of articles, which is far too many to label. Moreover, most of them would be irrelevant, i.e., they would consist of claims completely unrelated to the articles. To deal with this problem, we select for labelling only a subset of pairs with a high possibility to be relevant. We used two selection methods during our labelling.

At first, we used ElasticSearch to select a subset of the article-claim pairs. More specifically, we used each claim in turn as a query to find matching articles. This returned a large set of articles along with the BM25 score for each article. We kept only articles with the score higher than the $\frac{2}{3}$ of the maximum score, i.e., the score associated with the first matched article. We then shuffled the resulting set of article-claim pairs and sampled two batches, each with 100 random pairs, i.e., 200 pairs in total. We split them among six annotators so that each pair was assigned to three annotators. The annotations were collected using spreadsheets: each annotator was assigned one sheet per batch, with each row describing a single article-claim pair. For each article-claim pair, the annotators were presented with the title of the article, the claim, article URL and the claim URL for information.

However, this selection method led to a significant class imbalance. Out of 197 article-claim pairs, where there was an agreement between the annotators, the claims were labelled as present only in $\sim$10\% of cases, which also limited the number of stance annotations. We also observed a relatively large number of ``Can't tell'' labels which were caused by several claims. These mostly too generic claims (e.g., ``There are more doctors'') were mistakenly matched with many articles. The former was addressed by using our proposed claim presence detection baseline (cf. Section \ref{methods:presence}) instead of the simple querying in ElasticSearch. To mitigate the latter, we manually filtered out these problematic claims from further labelling.

We also switched from spreadsheets to a custom-made web-based annotation application, suitable also for mobile devices, which enabled us to reach a wider range of annotators. The application streamlined the annotation process and the article-claim pairs distribution to the annotators. The article-claim pairs were served to annotators until a match of at least two annotators was achieved in the values of claim presence as well as the article stance. Pairs with at least one label, but where no consensus had been achieved yet, were served to the annotators with a higher priority to keep the ``unfinished'' pair labels to a minimum.

\begin{figure}
    \centering
    \includegraphics[width=0.52\linewidth]{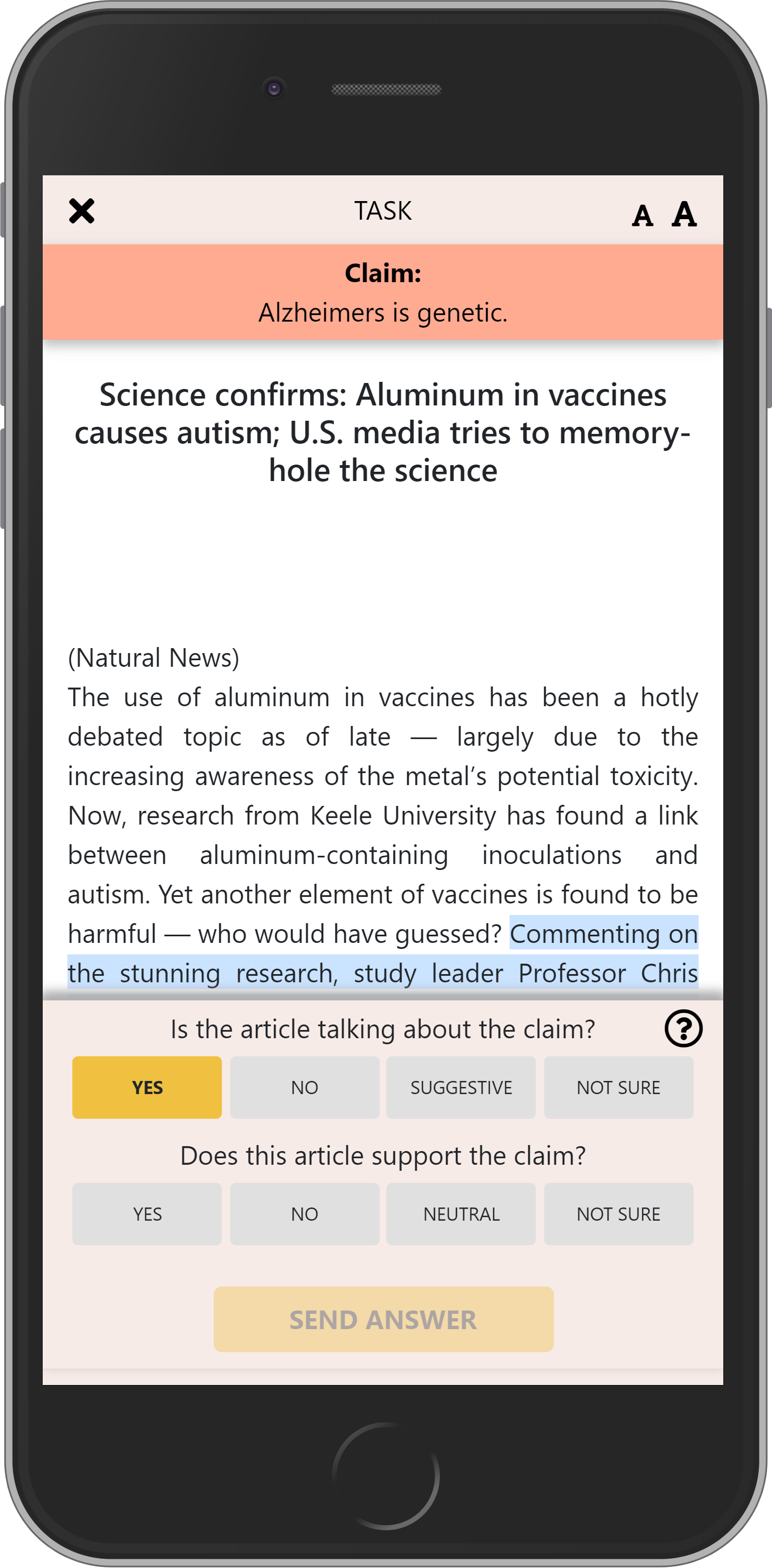}
    \caption{The mobile interface of the annotation application used in the later stage of article-claim pairs labelling. The annotators were presented with an article, a claim at the top, highlighted most similar sentence and buttons for selecting claim presence and article stance labels.}
    \label{fig:annotation_app}
\end{figure}

Each article-claim pair was presented in the application as shown in Figure~\ref{fig:annotation_app}. The claim was presented at the top, visually separated from the rest of the presented content. Underneath the claim, the title of the article, followed by its formatted body, was presented to the annotators. On the bottom, the annotators were presented with buttons for assigning the claim presence label and---if the annotators chose that the claim is present in the article---also the article stance label. As the articles were long and often dealt with multiple claims at the same time, we used a supportive text highlighting feature: the application highlighted sentences in the article that were most similar to the claim. The similarity was determined by cosine similarity between a sentence embedding representation of the given claim and the sentences of the article. Using this approach, we collected additional 376 article-claim pair labels from 28 annotators.

The collection of labels was also distributed in time. First {\numberOfManualLabelsSampleOne} article-claim pairs (denoted as \emph{Sample 1} in sections below) were annotated in 2019 and early 2020; since this was before the onset of the COVID-19 pandemic, this sample does not contain any claims or articles pertinent to it. The remaining {\numberOfManualLabelsSampleTwo} pairs (denoted as \emph{Sample 2} in sections below) were annotated in June 2021, thus capturing also narratives spread in that time.

\section{Dataset description}
\label{dataset}

\subsection{Descriptive analysis of raw data}
\label{dataset:raw}

The dataset consists of medical news articles/blogs and fact-checked claims in English language. However, the Monant platform, which was used to collect the dataset and makes it accessible via an API endpoint, also collects articles from other domains (e.g., politics or general news) and in other languages (currently mostly in Slovak and Czech). Out of approx. {\numberOfAllArticlesApprox} unique news articles/blogs from {\numberOfAllArticleSources} sources, there are {\numberOfArticlesApprox} English medical articles from {\numberOfArticleSources} sources.\footnote{The content of this section and section~\ref{methods:predictions} is based on the dataset's analysis published at: \url{https://github.com/kinit-sk/medical-misinformation-dataset/}. To make the analysis replicable, it uses a ``freeze time'' set to {\freezeTime}. As a result, only the data, that were present in the Monant platform up to this date, are considered.} Out of approx. {\numberOfAllFactchekingArticlesApprox} fact-checking articles extracted from {\numberOfAllFactchekingArticlesSources} fact-checking sites, there are {\numberOfClaimsApprox} fact-checked medical claims from {\numberOfClaimSources} fact-checking sites. And out of approx. {\numberOfAllDiscussionPostsApprox} discussion posts (related to {\numberOfAllDiscussionPostArticlesApprox} articles), there are {\numberOfDiscussionPostsApprox} discussions posts attached to English medical articles. In the following analyses, we focus specifically on English medical data contained in the provided dataset.

The dataset provides a rich set of features about each article. Besides an article's URL, title, textual body, and attached multimedia, it also contains information about article's authors, category, tags, and references. In addition, we collect (in regular intervals) the users' feedback on Facebook (i.e., the number of likes or shares) for each news article. In some cases, the posts from the attached discussions are available as well.

For {\numberOfArticleSourcesWithReliabilityAnnnotation} sources, we have an explicit source reliability (credibility) label (cf. Section~\ref{methodology:collection} for more details): {\numberOfArticleSourcesReliable} sources are considered to be reliable sources, {\numberOfArticleSourcesUnreliable} sources are considered to be unreliable. Out of all medical articles, {\numberOfArticlesFromReliableSourcesPercentageApprox} were collected from reliable sources, {\numberOfArticlesFromUnreliableSourcesPercentageApprox} from unreliable sources, and only {\numberOfArticlesFromSourcesWithoutReliabilityPercentageApprox} articles are from the sources without any reliability label.

\begin{figure}
   \centering
   \includegraphics[width=0.77\linewidth]{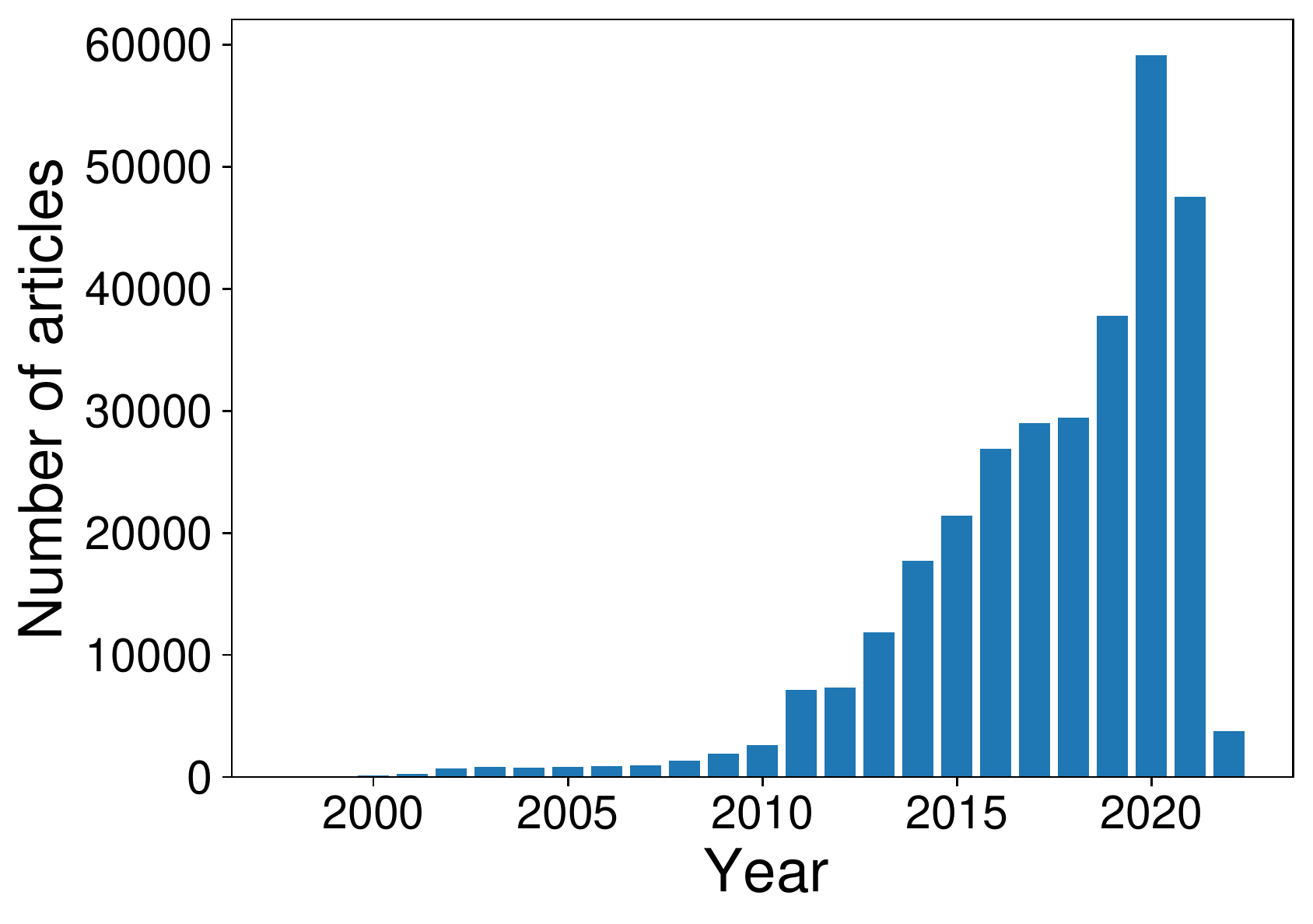}
   \caption{Number of collected medical articles in our dataset according to their publication year.}
   \label{fig:articles_timeline}
\end{figure}

Wherever possible, we collected all articles published by a given source. Consequently, some of the articles in the dataset were published as soon as 1998. Nevertheless, the majority of the collected news articles were published between years 2010--2021 as shown in Figure~\ref{fig:articles_timeline}. We can see an increasing trend in the number of medical news articles, with a significant increase in the last three years (the extreme rise in year 2020 can be explained by the onset of the COVID-19 pandemic).

\begin{figure}
   \centering
    \includegraphics[width=0.77\linewidth]{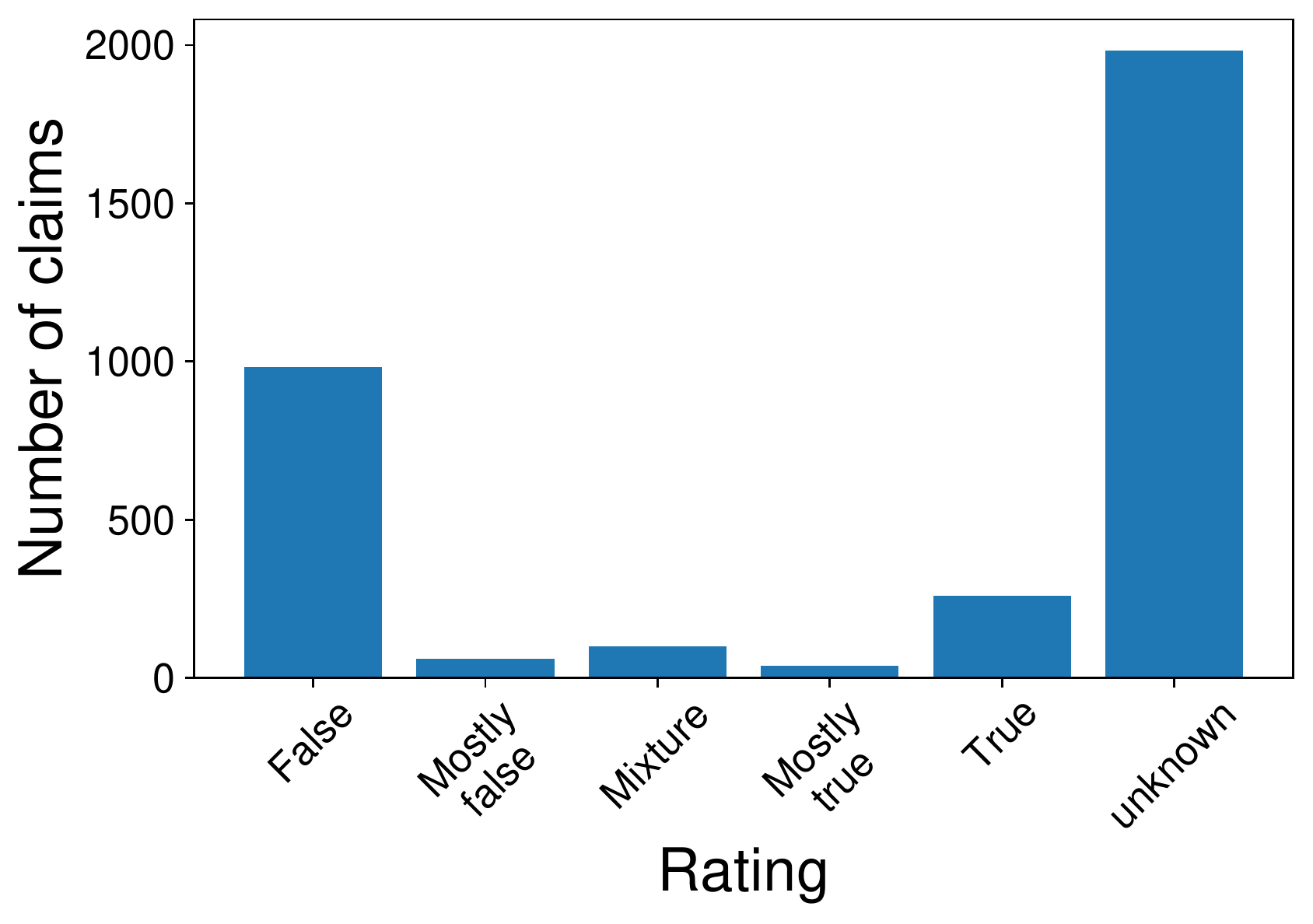}
    \caption{Number of medical claims in our dataset according to their veracity rating.}
    \label{fig:claim_ratings}
\end{figure}

Figure~\ref{fig:claim_ratings} shows the distribution of veracity ratings of the fact-checked medical claims contained in the dataset. {\numberOfFalseClaims} were evaluated as false, {\numberOfMostlyFalseClaims} as mostly false, {\numberOfMixtureClaims} as mixture, {\numberOfMostlyTrueClaims} as mostly true, and {\numberOfTrueClaims} as true. The rating of a significant number of claims (originating mostly from MetaFact.io, cf. Section~\ref{methodology:collection}) is currently unknown. 

\subsection{Labelled dataset}
\label{dataset:labelled}

The dataset contains {\numberOfManualLabels} article-claim pairs labelled by human annotators. There are {\numberOfManualClaimPresentLabels} pairs annotated with positive claim presence labels, out of which {\numberOfManualStanceLabels} also have article stance labels. The overall distribution of the claim presence and article stance labels is shown in Table~\ref{tab:distribution_of_labels}. It also shows distributions for individual Samples 1 and 2. As we can see, while there is a balance between present and not present labels in \emph{Sample 1} as well as overall, \emph{Sample 2} is skewed towards present labels. As to the article stance, most articles support the matched claims. There is a lack of ``Neutral'' stance labels in our dataset, i.e., of articles that would present both sides of the argument. This can make it difficult for models trained on this data to correctly classify this stance class.

Besides the labels from human annotators, the dataset also contains approx. {\numberOfClaimMappingsApprox} article-claim pairs with labels \textit{predicted} by our proposed baselines. Their analysis is provided in Section~\ref{methods:predictions}. 

\begin{table}
   \centering
   \caption{Distribution of claim presence and article stance labels in the dataset.}
   \label{tab:distribution_of_labels}
   \begin{tabular}{@{}lllllll@{}}
      \toprule
       & \emph{Sample 1} & \emph{Sample 2} & Overall    \\ \midrule
      \multicolumn{4}{c}{\textbf{Claim presence}} \\
      Present (incl. Suggestive) & {\numberOfManualClaimPresentLabelsSampleOne} ({\percentageOfManualClaimPresentLabelsSampleOne}) & {\numberOfManualClaimPresentLabelsSampleTwo} ({\percentageOfManualClaimPresentLabelsSampleTwo}) & {\numberOfManualClaimPresentLabels} ({\percentageOfManualClaimPresentLabels}) \\
      Not present & {\numberOfManualClaimNotPresentLabelsSampleOne} ({\percentageOfManualClaimNotPresentLabelsSampleOne}) & {\numberOfManualClaimNotPresentLabelsSampleTwo} ({\percentageOfManualClaimNotPresentLabelsSampleTwo})  & {\numberOfManualClaimNotPresentLabels} ({\percentageOfManualClaimNotPresentLabels}) \\ 
      \textit{Total} & {\numberOfManualLabelsSampleOne} & {\numberOfManualLabelsSampleTwo} & {\numberOfManualLabels} \\ \hline
      
      \multicolumn{4}{c}{\textbf{Article stance}} \\
      Supporting & {\numberOfManualStanceSupportingLabelsSampleOne} ({\percentageOfManualStanceSupportingLabelsSampleOne}) & {\numberOfManualStanceSupportingLabelsSampleTwo} ({\percentageOfManualStanceSupportingLabelsSampleTwo}) & {\numberOfManualStanceSupportingLabels} ({\percentageOfManualStanceSupportingLabels})  \\
      Contradicting & {\numberOfManualStanceContradictingLabelsSampleOne} ({\percentageOfManualStanceContradictingLabelsSampleOne}) & {\numberOfManualStanceContradictingLabelsSampleTwo} ({\percentageOfManualStanceContradictingLabelsSampleTwo}) & {\numberOfManualStanceContradictingLabels} ({\percentageOfManualStanceContradictingLabels}) \\
      Neutral & {\numberOfManualStanceNeutralLabelsSampleOne} ({\percentageOfManualStanceNeutralLabelsSampleOne})   & {\numberOfManualStanceNeutralLabelsSampleTwo} ({\percentageOfManualStanceNeutralLabelsSampleTwo})   & {\numberOfManualStanceNeutralLabels} ({\percentageOfManualStanceNeutralLabels}) \\
      \textit{Total} & {\numberOfManualStanceLabelsSampleOne} & {\numberOfManualStanceLabelsSampleTwo} & {\numberOfManualStanceLabels} \\ \bottomrule
   \end{tabular}
\end{table}

\subsection{Downstream tasks}
\label{dataset:tasks}

The collected dataset can support a range of fact-checking and misinformation-related tasks. Its main intended use is for training and evaluation of machine learning methods for the tasks of \emph{claim presence detection} and \emph{article stance classification}. The former can be considered a claim-oriented document retrieval problem, i.e., given a fact-checked claim, all documents, where it is present, are retrieved; or, alternatively, as previously fact-checked claims detection, i.e., given an unverified piece of text or claim, all relevant previously fact-checked claims are retrieved~\cite{Shaar2021clef}. The latter is a classification problem; the aim is to detect stance (position) of the author of an input piece of text towards a specified target~\cite{Kucuk2021stance}.

Since the dataset contains articles from a number of reliable and unreliable sources, it could be used for the \emph{misinformation characterisation} task, i.e., for analyses of characteristics of articles (how they are written) similar to~\cite{dhoju2019}: what topics they cover and how these topics evolve over time. The mapping of articles to fact-checked claims provides a straightforward grouping of the articles based on the misinformation they are related to. 

The misinformation sources often create inter-connected networks which spread and amplify the false information~\cite{Hrckova2021quantitative}. Since the dataset contains full-texts of the articles, it supports the task of \emph{misinformation spreading/diffusion analysis}. For example, it is possible to analyse linking patterns between the sources, search for content that is similar or even taken over from other sources, etc. Having a publication date of the articles, it is also possible to analyse where the misinformation first appeared and when (how fast) it was taken up by other sources. This is especially relevant with respect to the spread of misinformation between countries and across languages. Since the data available in Monant via an API endpoint also contain non-English sources (at this moment Slovak and Czech), it can be used to develop and test multilingual methods and analyse spreading patterns from English-language sources to other languages and (possibly) vice-versa.

Besides text, the dataset contains other modalities, such as image URLs, article and source metadata, etc. These can be all utilised to develop \emph{multimodal detection methods}. Lastly, the dataset can also be used for the task of \emph{source credibility identification} by utilising the existing source credibility labels and extracting a range of credibility indicators from the articles and available metadata, such as polarity of the articles, use of references, use of authors, etc.

\subsection{Ethical considerations}
The dataset was collected and is published for research purposes only.\footnote{To ensure that it will be used only for research, the access to the dataset is given upon request, in which the researchers give explicit consent with the usage conditions.} We collected only publicly available content of news articles/blogs. The dataset contains identities of authors of the articles if they were stated in the original source; we left this information, since the presence of an author's name can be a strong credibility indicator. However, we anonymised the identities of the authors of discussion posts included in the dataset.

The main identified ethical issue related to the presented dataset lies in the risk of mislabelling of an article as supporting a false fact-checked claim and, to a lesser extent, in mislabelling an article as not containing a false claim or not supporting it when it actually does. To minimise these risks, we developed our labelling methodology as described in Section~\ref{methodology:labelling} and require an agreement of at least two independent annotators to assign a claim presence or article stance label to an article. It is also worth noting that we do not label an article as a whole as false or true. Nevertheless, we provide partial article-claim pair veracities based on the combination of claim presence and article stance labels (cf. Section~\ref{methods:predictions}).

As to the veracity labels of the fact-checked claims and the credibility (reliability) labels of the articles' sources, we take these from the fact-checking sites and external listings such as Media Bias/Fact Check as they are and refer to their methodologies for more details on how they were established.

Lastly, the dataset also contains automatically predicted labels of claim presence and article stance using our baselines described in the next section. These methods have their limitations and work with certain accuracy as reported in this paper. This should be taken into account when interpreting them. 

The means for reporting considerable mistakes in the raw data and manual labels are described in the accompanying repository.\footnote{\url{https://github.com/kinit-sk/medical-misinformation-dataset/}}

\section{Claim presence and article stance baselines and analysis}
\label{methods}

\subsection{Evaluation of claim presence baselines}
\label{methods:presence}

\begin{table*}[ht!]
\centering
\caption{Precision, recall and F1-score of the claim presence detection baselines are evaluated on the whole manually labelled dataset. Accuracy is computed individually on the \emph{Sample 1} and \emph{Sample 2}, which were collected and annotated in 2019 and in 2021 respectively, as well as on the dataset as a whole.}
\label{tab:presence_methods_stats}
\begin{tabular}{@{}lllllll|lll@{}}
\toprule
             & \multicolumn{3}{l}{Present} & \multicolumn{3}{l|}{Not present} & \multirow{2}{*}{\emph{S1} Acc.} & \multirow{2}{*}{\emph{S2} Acc.} & \multirow{2}{*}{Overall Acc.} \\
             & Precision & Recall & F1-score & Precision & Recall & F1-score &  &  &  \\ \midrule
IR method    &  0.81  & 0.40  &  0.53  & 0.54  &  0.89  &  0.67  &  0.66  &  0.46  &  0.62  \\
SE method    &  0.79  & 0.43  &  0.56  & 0.55  &  0.86  &  0.67  &  0.65  &  0.53  &  0.62  \\
IRSE method  &  \textbf{0.91}  &  \textbf{0.45}  &  \textbf{0.6}  &  \textbf{0.58}  &  \textbf{0.95}  &  \textbf{0.72}  &  \textbf{0.71}  &  \textbf{0.56}  &  \textbf{0.67} \\ \bottomrule
\end{tabular}
\end{table*}

We provide evaluation of three claim presence detection baselines and compare their performance on the whole manually labelled dataset using only \textit{Not present} and \textit{Present} (which includes also \textit{Suggestive}) classes (see Table~\ref{tab:distribution_of_labels} for their distribution):

\begin{itemize}
    \item \textit{Information retrieval (IR method)} -- For any given claim and any given article, the claim presence is determined by the IR method as follows: First, 1-, 2- and 3-grams are extracted from the given claim. Each n-gram is assigned a TF-IDF score where TF is calculated within the claim and IDF based on the whole corpus of articles. Next, we match n-grams to the sentences of an article. If an n-gram contains medical terms, their synonyms are also allowed when matching sentences. Medical terms are identified using the Academic Vocabulary List.\footnote{\url{https://www.academicvocabulary.info/x.asp}} Synonyms to these terms are retrieved by comparing similarity of their word vectors using fastText\footnote{\url{https://fasttext.cc}} pre-trained on Wikipedia articles. The scores of n-grams, for which there is an article sentence containing all their terms, are summed up and normalized by the sum of all n-gram scores. We do this separately for 1-, 2-, and 3-grams and compute the final presence score as their average. The claim is classified as present in a given article if the final computed score is above a defined threshold.

    \item \textit{Sentence embedding similarity (SE method)} -- This method calculates a presence score based on sentence embeddings (using Universal Sentence Encoder~\cite{cer2018universal}, model v4) extracted from article sentences and a claim. The score is an average of two similarity comparisons: 1) cosine similarity between an article title and a claim, and 2) average cosine similarity between a claim and 5 article sentences the most similar to the claim. The claim is classified as present in a given article if the final computed score is above a defined threshold.

    \item \textit{Combined IRSE method} -- This method, first introduced in~\cite{pecher2020fireant}, works the same as the IR method with few important distinctions: First, the score of each matched n-gram is computed as a product of its TF-IDF score (IR method) and the cosine similarity between the embedding of an article sentence, which contains all terms of the given n-gram, and the claim embedding (SE method). Second, to make the comparison more efficient, only sentences with similarity above a certain threshold are considered. This threshold is computed as an average of cosine similarity between the claim and an article title embeddings and cosine similarity between the claim and the \textit{K} most similar sentences.
\end{itemize}

\subsubsection{Results of Claim Presence Detection}

All three baselines required a choice of a threshold to make the claim presence decisions based on the computed presence scores. We chose the threshold values so that recall of the methods on the positive (i.e., \textit{Present}) class would roughly be the same (around 0.4). This way, we can compare the methods working under the same requirement for the proportion of relevant items to be selected. The resulting thresholds for the \emph{IR}, \emph{SE}, and \emph{IRSE} methods were 0.5, 0.5, and 0.45 respectively. The \emph{IRSE} method also contains a prefiltering threshold. Our experiments showed that setting its value to 0.25 enabled it to discard a large number of potential mappings without affecting the overall performance of the method.

\begin{figure}
    \centering
    \includegraphics[width=0.4\textwidth]{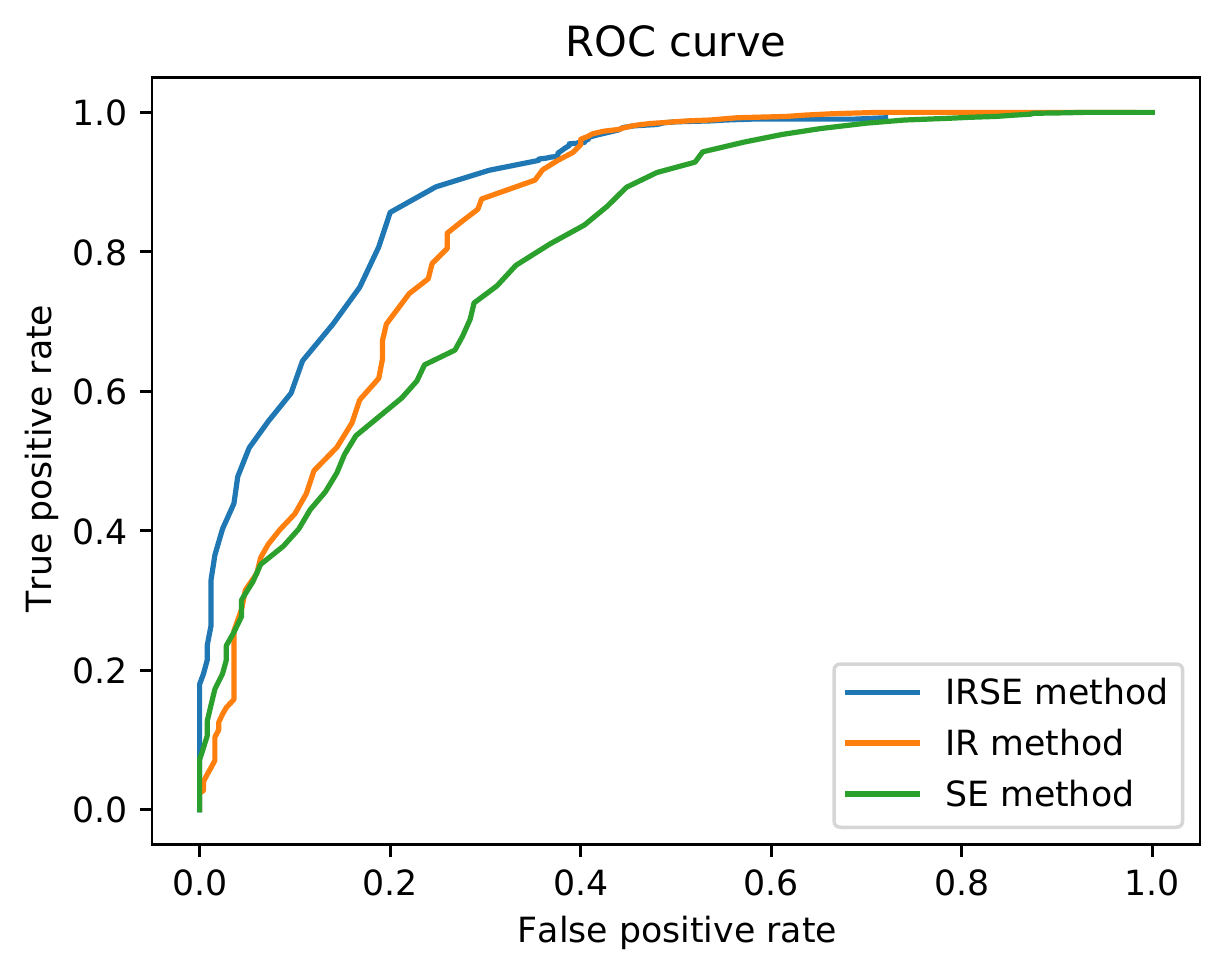}
    \caption{
        ROC curve showing relation between true positive rate and false positive rate of the evaluated baseline methods. The \emph{IRSE} method outperforms both the \emph{IR} and \emph{SE} methods by achieving lower false positive rate at most evaluated true positive rates.
    }
    \label{fig:roc}
\end{figure}

The results of the baselines on our labelled dataset are shown in Table~\ref{tab:presence_methods_stats}. Although the \emph{IR} and \emph{SE} methods achieved similar results, we can see that the \emph{IRSE} method outperformed both suggesting the utility of their combination. This is also confirmed by Figure~\ref{fig:roc}, which illustrates a relation between true positive rate and false positive rate of the baselines using ROC (receiver operating characteristic) curve. The \emph{IRSE} method retains lower false positive rate with increasing true positive rate than both the \emph{IR} and \emph{SE} methods. Out of them, the \emph{IR} method performs better with lower false positive rate than the \emph{SE} method.

We also evaluated the baselines individually for Samples 1 and 2 (see Table~\ref{tab:presence_methods_stats}). Although the \emph{IRSE} method retains the highest accuracy, the accuracy drops for all methods in \emph{Sample 2} compared to \emph{Sample 1}. Manual inspection of the errors made by the IRSE method revealed that the decrease cannot be explained by a domain shift due to COVID-19. Most commonly, the errors were due to the claim presence method neglecting some information in claims and mapping them to articles that were related but did not discuss that specific case. For instance, for claim ``Omega-3 fatty acids decrease triglycerides'', we observed results that discussed other effects of Omega-3 fatty acids that did not relate to triglycerides. To handle such cases, a more strict threshold could be used.

\subsection{Evaluation of article stance baselines}
\label{methods:stance}

To evaluate article stance classification baselines, we utilize \emph{Sample~1} with {\numberOfManualStanceLabelsSampleOne} pairs as training set and \emph{Sample~2} with {\numberOfManualStanceLabelsSampleTwo} pairs as testing set (see Table~\ref{tab:distribution_of_labels} for the distribution of classes in both samples). In both cases, we do not consider the \textit{Not present} class, as it is not relevant for stance classification. In addition to the manually labelled Monant data, we also utilise the Fake News Challenge (FNC) dataset\footnote{\url{http://www.fakenewschallenge.org/}}. Similarly, we drop the class denoting that an article is unrelated to the claim. This leaves us with $\sim$20,450 samples with the following distribution: 27.24\% \textit{Supporting}, 7.5\% \textit{Contradicting}, and 65.26\% \textit{Neutral}.

We compare the performance of several baselines. The first group of baselines present the best models from the FNC:

\begin{itemize}
    \item \textit{Talos}\footnote{\url{https://blog.talosintelligence.com/2017/06/talos-fake-news-challenge.html}} -- an ensemble of a decision tree and a convolutional neural network, where the final decision is obtained by simple 50/50 voting. This approach uses both hand-crafted features in the decision tree and the word embeddings in both the decision tree and the neural network.
    
    \item \textit{Athene}~\cite{hanselowski2018athene} -- an ensemble of multiple multi-layer perceptrons. The final decision is obtained by hard voting between them. All models use the same set of hand-crafted features, with only difference being their random initialisation.
    
    \item \textit{Athene-ext}~\cite{hanselowski2018athene} -- an extension of the \textit{Athene} approach, developed after the analysis of various models in the challenge, designed to overcome the observed problems. A single stacked LSTM is used with the best subset of the hand-crafted features, as determined by an ablation study.
    
    \item \textit{UCL}~\cite{riedel2017FNCbaseline} -- a simple multi-layer perceptron which uses TF-IDF scores of the claim and the article and the similarity between them.
\end{itemize}

Since the challenge took place already in 2017, these models can no longer be considered state of the art, but they nevertheless represent a relatively wide range of approaches utilising both hand-crafted, but also automatically extracted features, which makes them interesting for benchmarking more novel models.

The second group comprises our proposed baseline methods which utilise CNN and LSTM combined with similarity and attention mechanism respectively to identify parts of articles relevant for stance classification:  

\begin{itemize}

    \item \textit{All Sentences CNN} -- the input to the CNN model is a claim followed by the first 100 sentences of the article, without any detection of their relevance. The articles with higher number of sentences are clipped and those with lower number of sentences are padded with zero vector. This network is meant for comparison purposes, to determine the effect of sentence relevance detection.

    \item \textit{Attention LSTM} -- this model uses an LSTM network for obtaining high-level representations for both the claim and the article body. An attention mechanism is applied on the high-level representations to identify important parts of the articles. Another LSTM layer is applied on the output of the attention layer. A dropout with rate of 0.4 is applied to prevent overfitting, followed by a dense layer and a softmax layer for classification.
    
    \item \textit{Similarity CNN} -- the input to the CNN model are the three most similar sentences to a given claim (based on the cosine similarity of their embedding representations) along with one previous and one following sentence for each. We use three different convolutional layers, the outputs of which are concatenated together. A dropout of rate 0.25 is applied before the convolutional layers and one with rate of 0.5 is applied on the concatenated output of the convolutions, to prevent overfitting. Finally, we apply a dense layer and a softmax layer for classification. 
\end{itemize}

For the last two proposed baseline models, we also employ transfer learning. We first train a general model using the FNC data and fine-tune it on the manually labelled Monant training data.

\subsubsection{Results of Article Stance Classification}

Table~\ref{tab:stance-classification} presents accuracy of the baseline methods on the two datasets. In case of FNC dataset, we evaluate the models using a test subset of the dataset, as it was originally released for the competition. In case of manually labelled Monant data, we perform two evaluations. First, we perform a 5-fold cross-validation on the training set (\emph{Sample 1}) and report the mean performance of the model, which is determined by running the cross-validation 10 times. Second, we evaluate models trained on the whole \emph{Sample 1} on the testing set (\emph{Sample 2}).

The results show that the models that utilise simple hand-crafted features struggle when dealing with a different dataset. This is evident in the \textit{Athene} and its extension. We can presume that the used hand-crafted features are too specific for the FNC data and do not generalise well to the Monant data. On the other hand, the models with automatic feature extraction, which include UCL baseline model, \textit{Talos}, and our proposed models, show a better performance and better generalisation. In addition, we can see that these models retain their accuracy even on \emph{Sample 2} which was collected later than the training data and could theoretically include data or concept drifts.

\begin{table}[t]
\centering
\caption{Comparison of the article stance classification baselines. The reported metric is accuracy calculated on test subset in case of FNC. In case of manually labelled data from Monant, we report accuracy both as a mean of 10 runs of 5-fold cross-validation on \emph{Sample 1}, as well as on testing samples (\emph{Sample 2}). The best performance is achieved by the similarity CNN with transfer learning.}
\label{tab:stance-classification}
\begin{tabular}{llll}
\toprule
                            &  FNC    & \emph{S1} & \emph{S2} \\ \midrule
Talos                       &  66.93  &  42.57  &  48.00 \\
Athene                      &  67.81  &  14.36  &  15.00 \\ 
Athene-ext                  &  69.00  &  19.31  &  10.00 \\ 
UCL                         &  65.76  &  37.13  &  47.00 \\ \hline
All Sentences CNN           &  64.91  &  40.54  &  57.00 \\
Attention LSTM              &  63.19  &  43.78  &  40.00 \\ 
Similarity CNN              &  65.57  &  56.76  &  63.00 \\ \hline
Attention LSTM -- transfer  &  64.79  &  61.83  &  65.00 \\ 
Similarity CNN -- transfer  &  \textbf{71.86} & \textbf{74.23} & \textbf{73.00} \\ \bottomrule
\end{tabular}
\end{table}

The results also suggest that the identification of relevant parts of the articles is necessary when dealing with longer articles. In case of FNC data, where the average length of article is $\sim$16 sentences, the performance increase is not as evident. This may be due to the specificity of the shorter articles, which mostly deal with a single claim, and therefore can be considered relevant as a whole for the classification. However, when investigating the articles from Monant, where the average article length is $\sim$55 sentences, the increase in performance observed in \textit{Attention LSTM} and \textit{Similarity CNN} as opposed to \textit{All Sentences CNN}, is noticeable. In such articles, the extraction of features from the whole article results in a lot of noise, which causes problems for the classification.

When comparing attention mechanism with the similar sentences extracted using cosine similarity, we found out that the former sometimes struggled to identify relevant parts of articles. It tended to focus solely on the sentences similar to a given claim, while the arguments regarding the claims were often present in the surrounding sentences instead. Since \textit{Similarity CNN} took also these surroundings into account, it achieved a better performance.

Lastly, the use of transfer learning contributed to a significant increase of performance on the Monant data, even though the discrepancy in the distribution of classes across the datasets was significant. When we were training LSTM networks using transfer learning, they often broke down and started predicting the most dominant class in the data. Even though the use of attention mechanism helped in this regard, CNNs proved to be more stable and reliable for generating good claim and article representations and therefore attained better performance.

\subsection{Descriptive analysis of predicted labels}
\label{methods:predictions}

We use the best-performing baselines, i.e., the \emph{IRSE} method and the \emph{Similarity CNN with transfer learning}, to predict claim presence and article stance labels for articles and fact-checked claims in the collected dataset; these are also part of the published data. In addition, we aggregate these labels into article-claim pair veracities as follows: If an article agrees with a claim, we assign the veracity of the claim to the article-claim pair. If an article contradicts a claim, we assign to the article-claim pair the veracity opposite to the claim veracity. Lastly, if an article has a neutral stance towards a claim, or the veracity of the claim is \textit{unknown}, the article-claim pair is evaluated to be \textit{unknown} as well. 

The predicted labels are less precise compared to the manual ones, but at the same time, they are available for a much larger number of articles. They are also more accurate than many commonly used heuristics (e.g., the ones derived solely from the reliability of an article's source). This makes them ideal to be used as (weak) labels for other misinformation detection methods (based on articles' content style or context) while accepting some noise introduced by the methods' inaccuracy in some cases.

In total, there are approx. {\numberOfClaimMappingsApprox} article-claim mappings labelled with positive claim presence labels\footnote{Note that the dataset also contains additional {\numberOfClaimPresenceMappingsNotPresent} mappings labelled with the claim not present label. This happens when a presence score is below the set threshold, but still achieves a meaningful value -- it allows dataset users to use another (lower) threshold to increase recall at the expense of precision.} and consequently, also with article stance annotations. Out of all {\numberOfArticlesApprox} articles, {\percentageOfArticlesMappedToClaims} are mapped to at least one claim. Out of all {\numberOfClaimsApprox} medical claims, {\percentageOfClaimsMappedToArticlesApprox} are mapped to at least one article. The majority of predicted claim presence labels are related to claims from MetaFact.io ({\numberOfClaimsMetafactioPercentageApprox}), followed by FullFact.org ({\numberOfClaimsFullfactPercentageApprox}), HealthFeedback.org ({\numberOfClaimsHealthfeedbackPercentageApprox}), the list of cancer-related claims created in~\cite{ghenai2018} ({\numberOfClaimsGhenaiPercentageApprox}), and Snopes.com ({\numberOfClaimsSnopesPercentageApprox}). Out of all predicted article stance labels, {\numberOfClaimStanceSupportingPercentageApprox} are supporting, {\numberOfClaimStanceNeutralPercentageApprox} are neutral, and {\numberOfClaimStanceContradictingPercentageApprox} contradicting.

The resulting article-claim pair veracity labels ({\numberOfArticleClaimVeracityApprox} in total) have the following distribution: {\numberOfArticleClaimVeracityFalsePercentageApprox} are classified as false, {\numberOfArticleClaimVeracityMostlyfalsePercentageApprox} as mostly false, {\numberOfArticleClaimVeracityMixturePercentageApprox} as mixture, {\numberOfArticleClaimVeracityMostlytruePercentageApprox} as mostly true, {\numberOfArticleClaimVeracityTruePercentageApprox} as true, and finally {\numberOfArticleClaimVeracityUnknownPercentageApprox} of article-claim pairs are labelled as unknown. A high number of article-claim pairs labelled as unknown is caused by the fact that {\numberOfUnknownClaimsPercentageApprox} of medical claims (mostly from MetaFact.io) have an unknown veracity.

Out of {\numberOfArticlesMappedToClaimsApprox} articles mapped to at least one claim, {\numberOfArticleClaimVeracityTrueConsistentArticlesPercentageApprox} are mapped only to true article-claim pair veracity labels, {\numberOfArticleClaimVeracityFalseConsistentArticlesPercentageApprox} only to false article-claim pair veracity labels, and finally, {\numberOfArticleClaimVeracityInconsistentArticlesPercentageApprox} of articles contain a mixture of true and false article-claim pair veracity labels. The remaining articles are associated only with one or several article-claim mappings with unknown veracity. 

Regarding the source credibility labels, {\numberOfArticleClaimVeracityUnreliablePercentageApprox} of article-claim pair veracity labels relate to articles which come from unreliable sources. Out of them, {\numberOfArticleClaimVeracityUnreliableFalsePercentageApprox} label article-claim pairs as false and {\numberOfArticleClaimVeracityUnreliableTruePercentageApprox} as true. {\numberOfArticleClaimVeracityReliablePercentageApprox} of article-claim pair veracity labels relate to articles which come from reliable sources; out of them, {\numberOfArticleClaimVeracityReliableFalsePercentageApprox} label article-claim pairs as false and {\numberOfArticleClaimVeracityReliableTruePercentageApprox} as true. Although further investigation is needed, we can see that more veracity annotations relate to articles from unreliable sources (even when we consider the distribution of articles from un/reliable sources in our dataset). However, it also suggests that the information on the sources' credibility (commonly used as a heuristic to label articles) is not sufficient and the articles need to be assessed by the claims they make.

\section{Conclusions and future work}
\label{conclusions}


In this paper, we introduced a labelled dataset of medical articles with mappings to fact-checked claims for training and evaluation of machine learning methods supporting the fact-checking process. Besides providing a static dump of the dataset, we also provide a programmatic access to continuously updated data in our Monant platform. The platform has already been maintained for over 2.5 years, collecting, updating, and annotating new data. The main supported tasks are claim presence detection and article stance classification, for which we provide manual labels, and which are essential for searching and checking whether a new article contains claims that have already been fact-checked. In addition, the dataset enables a range of other tasks, such as misinformation characterisation studies, studies of misinformation diffusion, source credibility classification, etc. Thus, the dataset can be useful for researchers interested in misinformation, automatised or ML-supported fact-checking as well as for NLP and IR community in general.

We also present results of claim presence and article stance baselines which are used to generate predicted labels mapping articles to fact-checked claims. While the former are based on combination of classical IR approaches and sentence similarity, the latter use more advanced neural networks approaches combined with transfer learning to compensate for the limited number of labelled samples (and class imbalance, especially w.r.t. the \textit{neutral} class). The baselines leave plenty of space for improvement, e.g., by applying state-of-the-art pre-trained language models based on transformers. Also, they currently work only for content in English language and are strictly limited to textual content (i.e., they cannot detect presence of a claim in an image, such as a screenshot from a social medium post, meme, etc.).

As future work, we plan to: 1) extend the dataset with content in other languages; 2) develop multilingual methods of claim presence and article stance; and 3) apply them in a range of tasks, such as detection of previously fact-checked claims, mapping of these claims to additional online content, and to automate audits of misinformation prevalence in social media recommender systems~\cite{tomlein_audit_2021}. Since the scarcity of manually labelled data will likely remain a problem, we will continue focusing on machine learning approaches that can utilise unlabelled or limited labelled data, such as meta learning or weakly supervised learning. Furthermore, we will seek more efficient ways of navigating the selection of examples to label (active learning), and ways of gathering and exploiting previous experience from other tasks as is the case of transfer and meta learning.

\begin{acks}
This work was partially supported by The Ministry of Education, Science, Research and Sport of the Slovak Republic under the Contract No. 0827/2021; by the Central European Digital Media Observatory (CEDMO), a project funded by the European Union under the Contract No. 2020-EU-IA-0267; and by TAILOR, a project funded by EU Horizon 2020 research and innovation programme under GA No. 952215.
\end{acks}

\bibliographystyle{ACM-Reference-Format}
\bibliography{references}

\end{document}

%% file: stats.tex
\newcommand{\freezeTime}{February 1, 2022}

\newcommand{\numberOfAllArticles}{885,403}
\newcommand{\numberOfAllArticlesApprox}{885k}
\newcommand{\numberOfAllArticleSources}{256}
\newcommand{\averageNumberOfArticlesPerSource}{3,458.61}

\newcommand{\numberOfAllFactchekingArticles}{9,633}
\newcommand{\numberOfAllFactchekingArticlesApprox}{10k}
\newcommand{\numberOfAllFactchekingArticlesSources}{17}

\newcommand{\numberOfAllDiscussionPosts}{778,947}
\newcommand{\numberOfAllDiscussionPostsApprox}{780k}
\newcommand{\numberOfAllDiscussionPostArticles}{47,849}
\newcommand{\numberOfAllDiscussionPostArticlesApprox}{48k}

\newcommand{\numberOfArticlesAllLangs}{355,417}
\newcommand{\numberOfArticles}{316,832}
\newcommand{\numberOfArticlesApprox}{317k}
\newcommand{\numberOfArticlesSlovak}{17,815}
\newcommand{\numberOfArticlesCzech}{20,770}

\newcommand{\numberOfArticleSources}{207}
\newcommand{\numberOfArticleSourcesWithReliabilityAnnnotation}{70}
\newcommand{\numberOfArticleSourcesReliable}{22}
\newcommand{\numberOfArticleSourcesUnreliable}{48}

\newcommand{\numberOfArticlesFromReliableSources}{123,227}
\newcommand{\numberOfArticlesFromReliableSourcesPercentage}{38.89\%}
\newcommand{\numberOfArticlesFromReliableSourcesPercentageApprox}{39\%}
\newcommand{\numberOfArticlesFromUnreliableSources}{177,197}
\newcommand{\numberOfArticlesFromUnreliableSourcesPercentage}{55.93\%}
\newcommand{\numberOfArticlesFromUnreliableSourcesPercentageApprox}{56\%}
\newcommand{\numberOfArticlesFromSourcesWithoutReliability}{16,408}
\newcommand{\numberOfArticlesFromSourcesWithoutReliabilityPercentage}{5.18\%}
\newcommand{\numberOfArticlesFromSourcesWithoutReliabilityPercentageApprox}{5\%}
\newcommand{\percentageOfArticlesWithReliabilityLabel}{95\%}

\newcommand{\numberOfClaims}{3,423}
\newcommand{\numberOfClaimsApprox}{3.5k}
\newcommand{\numberOfClaimSources}{7}

\newcommand{\numberOfFalseClaims}{983}
\newcommand{\numberOfMostlyFalseClaims}{60}
\newcommand{\numberOfMixtureClaims}{100}
\newcommand{\numberOfMostlyTrueClaims}{39}
\newcommand{\numberOfTrueClaims}{259}
\newcommand{\numberOfUnknownClaims}{1,982}
\newcommand{\numberOfUnknownClaimsPercentage}{57.9\%}
\newcommand{\numberOfUnknownClaimsPercentageApprox}{58\%}

\newcommand{\numberOfDiscussionPosts}{710,993}
\newcommand{\numberOfDiscussionPostsApprox}{711k}

\newcommand{\numberOfManualClaimPresentLabelsSampleOne}{222}
\newcommand{\percentageOfManualClaimPresentLabelsSampleOne}{51\%}
\newcommand{\numberOfManualClaimNotPresentLabelsSampleOne}{217}
\newcommand{\percentageOfManualClaimNotPresentLabelsSampleOne}{49\%}
\newcommand{\numberOfManualLabelsSampleOne}{439}

\newcommand{\numberOfManualClaimPresentLabelsSampleTwo}{101}
\newcommand{\percentageOfManualClaimPresentLabelsSampleTwo}{75\%}
\newcommand{\numberOfManualClaimNotPresentLabelsSampleTwo}{33}
\newcommand{\percentageOfManualClaimNotPresentLabelsSampleTwo}{25\%}
\newcommand{\numberOfManualLabelsSampleTwo}{134}

\newcommand{\numberOfManualClaimPresentLabels}{323}
\newcommand{\percentageOfManualClaimPresentLabels}{56\%}
\newcommand{\numberOfManualClaimNotPresentLabels}{250}
\newcommand{\percentageOfManualClaimNotPresentLabels}{44\%}
\newcommand{\numberOfManualLabels}{573}

\newcommand{\numberOfManualStanceSupportingLabelsSampleOne}{129}
\newcommand{\percentageOfManualStanceSupportingLabelsSampleOne}{61\%}
\newcommand{\numberOfManualStanceContradictingLabelsSampleOne}{62}
\newcommand{\percentageOfManualStanceContradictingLabelsSampleOne}{30\%}
\newcommand{\numberOfManualStanceNeutralLabelsSampleOne}{19}
\newcommand{\percentageOfManualStanceNeutralLabelsSampleOne}{9\%}
\newcommand{\numberOfManualStanceLabelsSampleOne}{210}

\newcommand{\numberOfManualStanceSupportingLabelsSampleTwo}{74}
\newcommand{\percentageOfManualStanceSupportingLabelsSampleTwo}{75\%}
\newcommand{\numberOfManualStanceContradictingLabelsSampleTwo}{24}
\newcommand{\percentageOfManualStanceContradictingLabelsSampleTwo}{24\%}
\newcommand{\numberOfManualStanceNeutralLabelsSampleTwo}{1}
\newcommand{\percentageOfManualStanceNeutralLabelsSampleTwo}{1\%}
\newcommand{\numberOfManualStanceLabelsSampleTwo}{99}

\newcommand{\numberOfManualStanceSupportingLabels}{203}
\newcommand{\percentageOfManualStanceSupportingLabels}{66\%}
\newcommand{\numberOfManualStanceContradictingLabels}{86}
\newcommand{\percentageOfManualStanceContradictingLabels}{28\%}
\newcommand{\numberOfManualStanceNeutralLabels}{20}
\newcommand{\percentageOfManualStanceNeutralLabels}{6\%}
\newcommand{\numberOfManualStanceLabels}{309}

\newcommand{\numberOfClaimMappingsApprox}{51k}
\newcommand{\numberOfClaimPresenceMappings}{50,953}
\newcommand{\numberOfClaimPresenceMappingsNotPresent}{366k}

\newcommand{\numberOfArticlesMappedToClaims}{34,850}
\newcommand{\numberOfArticlesMappedToClaimsApprox}{35k}
\newcommand{\percentageOfArticlesMappedToClaims}{11\%}
\newcommand{\numberOfClaimsMappedToArticles}{1,193}
\newcommand{\percentageOfClaimsMappedToArticles}{34.85\%}
\newcommand{\percentageOfClaimsMappedToArticlesApprox}{35\%}

\newcommand{\numberOfClaimsMetafactio}{33,950}
\newcommand{\numberOfClaimsMetafactioPercentage}{66.63\%}
\newcommand{\numberOfClaimsMetafactioPercentageApprox}{66.6\%}
\newcommand{\numberOfClaimsGhenai}{1,906}
\newcommand{\numberOfClaimsGhenaiPercentage}{3.74\%}
\newcommand{\numberOfClaimsGhenaiPercentageApprox}{3.7\%}
\newcommand{\numberOfClaimsHealthfeedback}{4,878}
\newcommand{\numberOfClaimsHealthfeedbackPercentage}{9.57\%}
\newcommand{\numberOfClaimsHealthfeedbackPercentageApprox}{9.6\%}
\newcommand{\numberOfClaimsFullfact}{9,418}
\newcommand{\numberOfClaimsFullfactPercentage}{18.48\%}
\newcommand{\numberOfClaimsFullfactPercentageApprox}{18.4\%}
\newcommand{\numberOfClaimsSnopes}{793}
\newcommand{\numberOfClaimsSnopesPercentage}{1.56\%}
\newcommand{\numberOfClaimsSnopesPercentageApprox}{1.6\%}

\newcommand{\numberOfClaimStance}{50,953}
\newcommand{\numberOfClaimStanceSupporting}{40,168}
\newcommand{\numberOfClaimStanceSupportingPercentage}{78.83\%}
\newcommand{\numberOfClaimStanceSupportingPercentageApprox}{79\%}
\newcommand{\numberOfClaimStanceNeutral}{2,065}
\newcommand{\numberOfClaimStanceNeutralPercentage}{4.05\%}
\newcommand{\numberOfClaimStanceNeutralPercentageApprox}{4\%}
\newcommand{\numberOfClaimStanceContradicting}{8,720}
\newcommand{\numberOfClaimStanceContradictingPercentage}{17.11\%}
\newcommand{\numberOfClaimStanceContradictingPercentageApprox}{17\%}

\newcommand{\numberOfArticleClaimVeracity}{50,953}
\newcommand{\numberOfArticleClaimVeracityApprox}{51k}
\newcommand{\numberOfArticleClaimVeracityFalse}{10,170}
\newcommand{\numberOfArticleClaimVeracityFalsePercentage}{19.96\%}
\newcommand{\numberOfArticleClaimVeracityFalsePercentageApprox}{20\%}
\newcommand{\numberOfArticleClaimVeracityMixture}{55}
\newcommand{\numberOfArticleClaimVeracityMixturePercentage}{0.11\%}
\newcommand{\numberOfArticleClaimVeracityMixturePercentageApprox}{0.1\%}
\newcommand{\numberOfArticleClaimVeracityMostlyfalse}{45}
\newcommand{\numberOfArticleClaimVeracityMostlyfalsePercentage}{0.09\%}
\newcommand{\numberOfArticleClaimVeracityMostlyfalsePercentageApprox}{0.1\%}
\newcommand{\numberOfArticleClaimVeracityMostlytrue}{41}
\newcommand{\numberOfArticleClaimVeracityMostlytruePercentage}{0.08\%}
\newcommand{\numberOfArticleClaimVeracityMostlytruePercentageApprox}{0.1\%}
\newcommand{\numberOfArticleClaimVeracityTrue}{8,814}
\newcommand{\numberOfArticleClaimVeracityTruePercentage}{17.30\%}
\newcommand{\numberOfArticleClaimVeracityTruePercentageApprox}{17\%}
\newcommand{\numberOfArticleClaimVeracityUnknown}{31,828}
\newcommand{\numberOfArticleClaimVeracityUnknownPercentage}{62.47\%}
\newcommand{\numberOfArticleClaimVeracityUnknownPercentageApprox}{62\%}

\newcommand{\numberOfArticleClaimVeracityTrueConsistentArticles}{7,348}
\newcommand{\numberOfArticleClaimVeracityTrueConsistentArticlesPercentage}{21.08\%}
\newcommand{\numberOfArticleClaimVeracityTrueConsistentArticlesPercentageApprox}{21\%}
\newcommand{\numberOfArticleClaimVeracityFalseConsistentArticles}{7,680}
\newcommand{\numberOfArticleClaimVeracityFalseConsistentArticlesPercentage}{22.04\%}
\newcommand{\numberOfArticleClaimVeracityFalseConsistentArticlesPercentageApprox}{22\%}
\newcommand{\numberOfArticleClaimVeracityInconsistentArticles}{971}
\newcommand{\numberOfArticleClaimVeracityInconsistentArticlesPercentage}{2.79\%}
\newcommand{\numberOfArticleClaimVeracityInconsistentArticlesPercentageApprox}{3\%}

\newcommand{\numberOfArticleClaimVeracityUnreliable}{35,094}
\newcommand{\numberOfArticleClaimVeracityUnreliablePercentage}{68.88\%}
\newcommand{\numberOfArticleClaimVeracityUnreliablePercentageApprox}{69\%}
\newcommand{\numberOfArticleClaimVeracityUnreliableFalse}{7,571}
\newcommand{\numberOfArticleClaimVeracityUnreliableFalsePercentage}{21.57\%}
\newcommand{\numberOfArticleClaimVeracityUnreliableFalsePercentageApprox}{22\%}
\newcommand{\numberOfArticleClaimVeracityUnreliableTrue}{5,794}
\newcommand{\numberOfArticleClaimVeracityUnreliableTruePercentage}{16.51\%}
\newcommand{\numberOfArticleClaimVeracityUnreliableTruePercentageApprox}{17\%}
\newcommand{\numberOfArticleClaimVeracityReliable}{12,815}
\newcommand{\numberOfArticleClaimVeracityReliablePercentage}{25.15\%}
\newcommand{\numberOfArticleClaimVeracityReliablePercentageApprox}{25\%}
\newcommand{\numberOfArticleClaimVeracityReliableFalse}{2,186}
\newcommand{\numberOfArticleClaimVeracityReliableFalsePercentage}{17.06\%}
\newcommand{\numberOfArticleClaimVeracityReliableFalsePercentageApprox}{17\%}
\newcommand{\numberOfArticleClaimVeracityReliableTrue}{2,549}
\newcommand{\numberOfArticleClaimVeracityReliableTruePercentage}{19.89\%}
\newcommand{\numberOfArticleClaimVeracityReliableTruePercentageApprox}{20\%}